\documentclass[sn-mathphys-num]{sn-jnl}

\usepackage{graphicx,subcaption}%
\usepackage{multirow}%
\usepackage{amsmath,amssymb,amsfonts}%
\usepackage{amsthm}%
\usepackage{mathrsfs}%
\usepackage[title]{appendix}%
\usepackage{xcolor}%
\usepackage{textcomp}%
\usepackage{manyfoot}%
\usepackage{booktabs}%
\usepackage{algorithm}%
\usepackage{algorithmicx}%
\usepackage{algpseudocode}%
\usepackage{listings}%

\usepackage{mathtools}
\usepackage{graphics}
\usepackage{float}
\usepackage{lipsum,afterpage,refcount}

\usepackage{xcolor}

\begin{document}

\title[Article Title]{Rethinking and Recomputing the Value of Machine Learning Models}


\author*[1]{\fnm{Burcu} \sur{Sayin}}\email{burcu.sayin@unitn.it}

\author[2]{\fnm{Jie} \sur{Yang}}\email{j.yang-3@tudelft.nl}

\author[2]{\fnm{Xinyue} \sur{Chen}}\email{xinyuechen223@gmail.com}

\author[1]{\fnm{Andrea} \sur{Passerini}}\email{andrea.passerini@unitn.it}

\author[1,3]{\fnm{Fabio} \sur{Casati}}\email{fabio.casati@servicenow.com}

\affil*[1]{\orgdiv{Department of Information Engineering and Computer Science}, \orgname{University of Trento}, \orgaddress{\street{Via Sommarive 9, Povo}, \city{Trento}, \postcode{38123}, \country{Italy}}}

\affil[2]{\orgdiv{Department of Software Technology}, \orgname{TU Delft}, \orgaddress{\street{Mekelweg 5}, \city{Delft}, \postcode{2628 XE},  \country{The Netherlands}}}

\affil[3]{\orgname{Servicenow}, \orgaddress{\street{Zurich}, \state{ZH}, \country{Switzerland}}}

\abstract{In this paper, we argue that the prevailing approach to training and evaluating machine learning models often fails to consider their real-world application within organizational or societal contexts, where they are intended to create beneficial value for people. We propose a shift in perspective, redefining model assessment and selection to emphasize integration into workflows that combine machine predictions with human expertise, particularly in scenarios requiring human intervention for low-confidence predictions. Traditional metrics like accuracy and f-score fail to capture the beneficial value of models in such hybrid settings. To address this, we introduce a simple yet theoretically sound ``value'' metric that incorporates task-specific costs for correct predictions, errors, and rejections, offering a practical framework for real-world evaluation. Through extensive experiments, we show that existing metrics fail to capture real-world needs, often leading to suboptimal choices in terms of value when used to rank classifiers. Furthermore, we emphasize the critical role of calibration in determining model value, showing that simple, well-calibrated models can often outperform more complex models that are challenging to calibrate.}




\maketitle

\section{Introduction}

\begin{figure}[b]
\centering
 \includegraphics[width=0.9\textwidth]{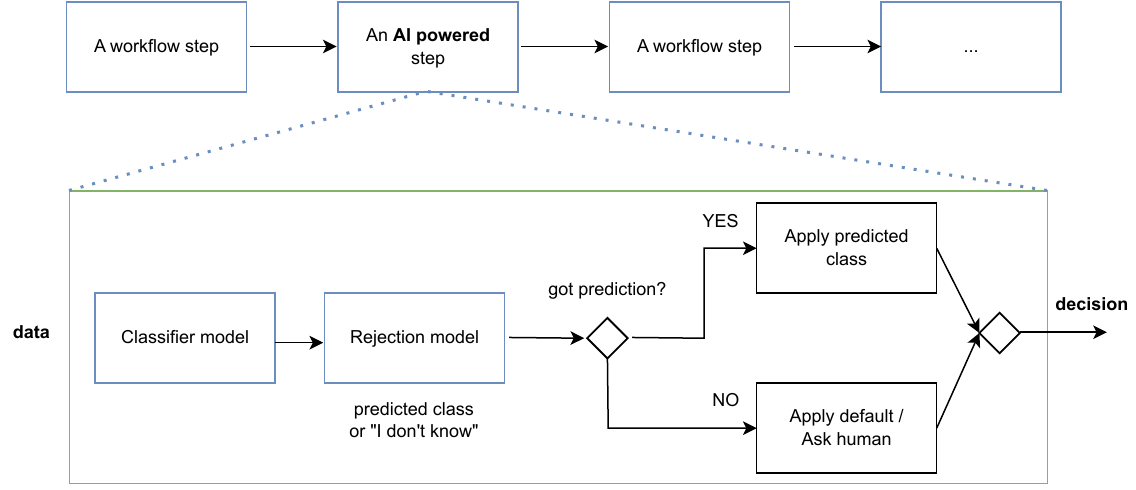}
\caption{A typical implementation of ML models into an ML solution workflow involves using a rejection function that filters predictions based on a confidence threshold. This approach generally assumes that the classifier is trained independently of the rejection logic. However, this is not a necessity—the classifier can be designed to be aware of the associated costs, which may make it less ``general" but more tailored to specific needs.}
\label{fig:wf}
\end{figure}

Recently, a few position papers~\cite{casatiwdhm21,Sayin2021HybridClassification,Sayin2021,burcuThesis2022,sayin2022rethinking,sayin2023ValueBasedHI,sayin2023ValueBasedAL} have challenged the underlying assumptions of quality in Machine Learning (ML), particularly the overemphasis on accuracy-based metrics and various measures of calibration errors (i.e. the difference between a model's predicted probabilities and the actual likelihood of its predictions being correct). At the heart of this stance, there are two observations: (i) ML models are almost always applied in hybrid human-machine settings, where the model can abstain or its prediction be rejected for insufficient confidence (i.e. the model's estimation of the correctness of its prediction) as in Figure \ref{fig:wf}, and (ii) the beneficial value of correct inferences, as well as the detrimental value of incorrect inferences and rejections, is determined by the use case, not by the model.

In our experience (see Section \ref{insights}), we have found that the majority of AI deployments in the enterprise consist of selective models or \textit{selective classifiers}~\cite{Geifman2017}, which is more of a rule than an exception. An example where this commonly occurs is in customer support requests, where the goal is to identify the customer's intent to trigger an automated request processing workflow if possible. Failing to comprehend the customer's intent and resorting to human agents is not ideal. However, it's even more problematic to misinterpret the customer's intent and guide them down the wrong path toward a resolution. This is why intent classifications are filtered based on prediction confidence. How ``good'' or ``useful'' a model is therefore depends on the beneficial value it brings when inserted in ML solution workflows (Figure \ref{fig:wf}). This beneficial value depends on how often the workflow rejects the predictions, on the correctness patterns of the predictions that are not rejected, and on the detrimental value of errors vs benefits of correct predictions. 
While in this paper we only marginally discuss the plethora of Large Language Models (LLMs), the problem is exactly the same if not worse: With generative AI the question of whether to show an answer or to withhold it is crucial and there are many things that can be wrong in an answer, the most common being hallucination. We add that the fact that some APIs do not reveal likelihood/confidence level makes model evaluation more difficult.

To some extent, all this is trivial. There is no inherent difficulty in developing use case-based value functions, selecting the best model from a set of well-performing models based on the value function, or evaluating a model's performance across multiple value functions. Moreover, one could contend that accuracy metrics are a sufficient substitute for evaluating model improvements in data science, or for selecting models to deploy in an AI platform designed to meet specific use cases. Thus, the practical approach would be to choose the model with the best accuracy or F1 score and enable users to filter out predictions with a confidence level lower than a set threshold. Accuracy and similar metrics are easy to comprehend and do not require us to determine parameters such as the ``cost of errors'', which can be difficult to estimate, especially when considering the use case.

In this paper, we show that this reasoning is wrong. If we accept that classifiers are mostly applied as selective models, then the method we use to measure, compare, and even train models must change. The implications of models being almost always applied as selective classifiers are often neglected in the literature, and this is also reflected in model leaderboards. We also show that the simplicity of not having to choose a cost parameter is an illusion: when we use accuracy to compare models, i) we do implicitly choose a cost parameter, often without realizing it, and ii) this implicitly selected cost is probably one of the worst choices possible: that of setting the relative cost of errors to zero. Despite being counter-intuitive, we show that accuracy is a quality metric that may be selected when the consequences of model errors are not critical. When a model is likely to be used across multiple use cases, relying solely on accuracy-based metrics can have significant implications. Overall, we show that:

\begin{itemize}
\item Universal metrics used for model evaluation are poor indicators of model value, potentially leading to incorrect decisions such as choosing models with {\em negative} value; 
\item Metrics designed to account for cost-sensitive errors are also inappropriate as they fail to consider the reject option;
\item Lack of calibration substantially affects model value, and poorly calibrated complex models can be outperformed by simple, decades-old models that are easier to calibrate;
\item Operating in an out-of-distribution setting further reduces the reliability of standard performance metrics.
\end{itemize}

It is worth underlining that the notion of value we introduce in this paper is not a radically different metric, but rather a combination of existing metrics, such as accuracy, detrimental value of errors and rejection rate, into a single measure accounting for the ``value'' of the predictor for a user. Importantly, the metric is normalized in such a way that a value of zero indicates a classifier that is completely useless, a negative value a classifier that is harmful (with respect to always ignoring it and resorting to the default path) and any value larger than zero indicates the gain that is obtained by using the classifier.

The remainder of this paper is structured as follows: in Section~\ref{relatedWork}, we review related works on our concept of model value. Then, in Section~\ref{sec:value}, we formalize this notion and introduce the rejection threshold maximizing value, along with its extension to the cost-sensitive setting where different errors have different costs. Section~\ref{sec:expWork} presents our experimental analysis comparing our value metric with standard performance measures, while Section~\ref{conclusion} offers our conclusions.
\section{Related work} \label{relatedWork}

\noindent\textbf{Selective classification.} Mimicking the typical use of ML models in many practical applications, a number of approaches rely on the combination of an ML model making an initial prediction and a human annotator taking over when the model's confidence is not high enough \cite{Law_hearth_cscw18}. Selective classifiers are specifically conceived for this use, by including a rejection mechanism to decide when to abstain from making a prediction. The literature on selective classifiers is extensive, encompassing a broad range of learning algorithms, including nearest-neighbor classifiers~\cite{Hellman1970NNwithReject}, SVM~\cite{Fumera2002SVMwithReject}, and neural networks~\cite{Cordella1995ImproveMLP,Stefano2000Reject,Geifman2017} (see Hendrickx et al.~\cite{Hendrickx2021} for a recent survey). The effectiveness of this solution is, however, heavily dependent on the reliability of machine confidence, which has shown to be very poor, especially for deep learning \cite{Balda2020,Guo2017CalibrationOfMNNs}.

\noindent\textbf{Classifier confidence.} To effectively use a classifier~\cite{Jiang2018ToTrust}, it is important to understand its properties and have confidence in its individual predictions. The literature proposes various confidence-based methods, including measuring the entropy of the softmax predictions~\cite{Teerapittayanon2017}, calculating trust scores based on the distance of samples to a calibration set~\cite{Jiang2018ToTrust}, determining a confidence threshold (via Shannon entropy~\cite{ShannonEntropy2001}, Gini coefficient~\cite{Bendel1989Gini}, or norm-based methods~\cite{AndrewNg2004Norms}) that maximizes coverage for a given accuracy~\cite{Bukowski2021}, and using semantics-preserving data transformation to estimate confidence~\cite{Bahat2020ClassificationCE}. Post-hoc recalibration is a popular strategy for improving classifier confidence, with techniques ranging from temperature scaling~\cite{Guo2017CalibrationOfMNNs} to Dirichlet calibration~\cite{Kull2019} (see a recent survey by Filho et al. ~\cite{calibration2021}). However, as we will show in our experimental evaluation (see Section~\ref{sec:calibration}), it's essential to complement these solutions with a proper value metric to assess the classifier's beneficial value in real-world applications.

\noindent\textbf{Cost-sensitive learning} addresses the challenge of training classifiers by considering the varying costs associated with different types of errors, particularly in scenarios with significant class imbalance~\cite{Elkan2001CostSensitiveLearning,Ling2010,ThaiNghe2010,Tu2020CostSensitive,pmlr-v139-charoenphakdee21a}. Existing work includes \cite{Tu2020CostSensitive}: (i) data-level approaches \cite{Ting1998,Zadrozny2003CostSensitive} where the class distribution of training data is balanced via sampling methods, and (ii) algorithm-level approaches, that use a thresholding scheme \cite{Chai2004TestcostSN,Domingos1999,Elkan2001CostSensitiveLearning,Ling2010,Sayin2021Survey,Sheng2006,Suri2022} to improve the prediction performance on the minority class (e.g. in binary classification, the threshold is set such that the prediction is 1 only if the expected cost associated with this prediction is lower than or equal to that of predicting 0). Although this line of work is closely related to our setting as it also considers the impact of errors on the downstream pipeline, it assumes that the classifier provides a prediction for every instance without any rejection mechanism. This assumption can significantly impact the evaluation of the resulting classifier's quality, as our experimental evaluation will demonstrate (see Section~\ref{sec:costsensitive_res}). Finally, \cite{pmlr-v139-charoenphakdee21a} introduces a novel approach to classification with rejection option by training an ensemble of cost-sensitive classifiers. In contrast, our goal is not to develop a novel cost-sensitive classifier. Instead, we aim to introduce a metric designed to evaluate such classifiers.

\noindent\textbf{Hybrid Human-AI systems} aim at solving classification problems with humans and machines~\cite{r2,r1,Raghu2019,WilderHK20}, but effectively combining human and machine intelligence has many challenges. For example, \textit{trust in humans} requires a deep understanding of how to design crowdsourcing tasks and model their complexity \cite{Ujwal2017,Qarout2018,Wu2017,Jie2016}, test and filter crowd workers \cite{Bragg2016}, aggregate results into a decision \cite{Han2020,Kamar_2012_combining,Evgeny2018,Li2013,Liu2013,Whitehill2009,Zhou2012}, improve the engagement \cite{Han2019,Han20192,Sihang2020}, or leverage crowds to learn features of ML models~\cite{flock_2015,Carlos_pattern}. Furthermore, \textit{the effective aggregation of human and machine decisions} \cite{Nagar2011,Nagar2012,Andres2022} depends on many factors, such as training, explaining, sustaining, interacting, and amplifying. The value metric is defined in the context of hybrid human-AI systems, where humans intervene whenever the AI defers a decision due to low confidence in its prediction. This metric accounts for the value of deferral, along with the impact of both correct and incorrect machine predictions, in assessing the overall value of the system. We believe that defining appropriate measures of the beneficial value of the joint human-machine system is a major prerequisite to keep research in the field on the right course.
\section{Measuring model ``value''}
\label{sec:value}

In this section, we formally define the notion of model ``value'', and show how threshold-based selective classifiers, by far the most popular class of classifiers in practical ML workflows, can be adjusted to maximize value. 

\subsection{The setting}
\begin{figure}[!htbp]
\centering
\includegraphics[width=0.95\textwidth]{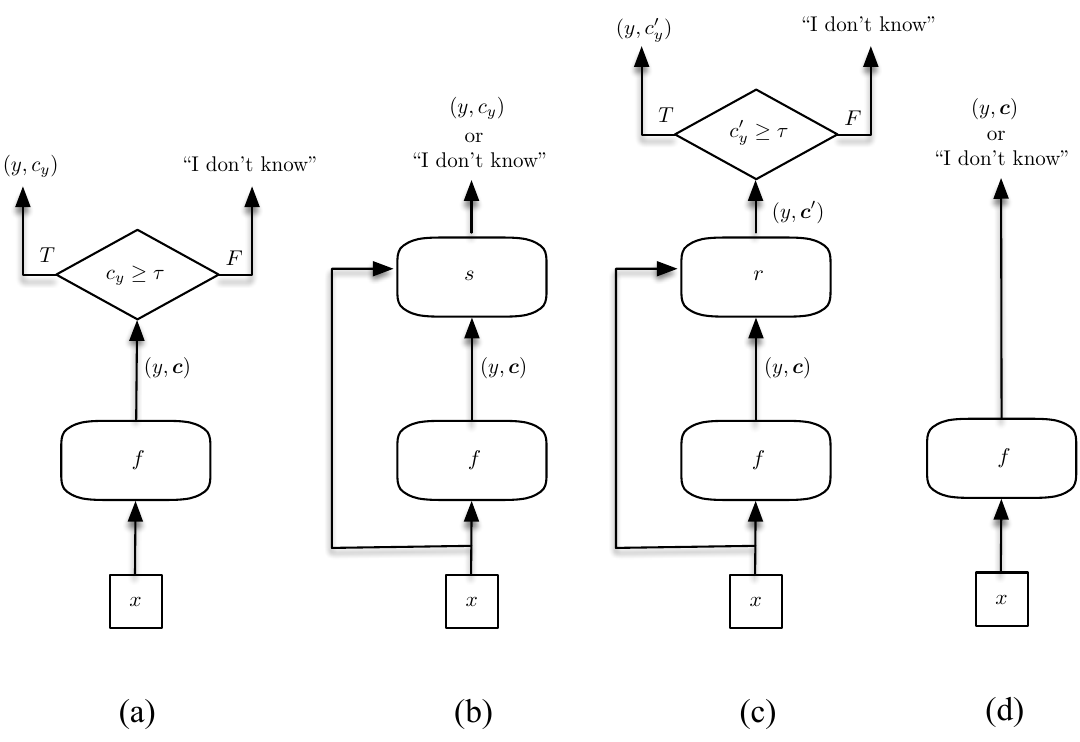}
\caption{Common approaches to selectivity in classification: (a) filtering predictions based on a confidence threshold, (b) employing an input-based selector model to decide on prediction acceptance, (c) using a confidence recalibrator followed by threshold-based filtering, or (d) incorporating built-in abstention with an `I don't know' class.}
\label{fig:options}
\end{figure}
Selective classifiers are ML models that generate output only when they are sufficiently confident in their prediction accuracy; otherwise, they abstain from making a decision, guided by a predefined rejection function. Selective classifiers can be implemented as follows:
\begin{enumerate}
\item[(a)] We take a model $f$ that outputs a prediction $y$ and a \textit{confidence} $c_y$ (or a vector $\mathbf{c}$ of confidence for a set of possible answers). Then, we filter the predictions to take only those above a certain confidence threshold (Fig.~\ref{fig:options}a).
\item[(b)] The model $f$ outputs predictions and confidence, but we apply a selector model $s$ that decides whether to accept the prediction or not, based on features of the input $x$ (Fig.~\ref{fig:options}b). 
\item[(c)] A hybrid of the two above cases is where the selector is a recalibrator $r$ that can either take as input the prediction and confidence measure (\textit{feature-agnostic} calibrator) or also the input features of $x$ and adjust the confidence vector (\textit{feature-aware} calibrator), typically applying threshold-based selection on the resulting confidence (Fig.~\ref{fig:options}c).
\item[(d)] The model $f$ is already trained to only output predictions that are ``good enough'' and includes an ``I don't know'' class (Fig.~\ref{fig:options}d).
\end{enumerate}

The first case is the most common, at least in our experience (see Section \ref{insights}). The second case is an extension and generalization of the first case in two ways: it can take features as input ($s$ can be trained as opposed to ``just'' being a formula), and it can filter based on any formula. It however requires some form of ``training'' or machine teaching, which is highly non-trivial. The recalibrator also typically requires some form of training. However, a feature-agnostic calibrator can be easily set up by post-hoc calibration strategies~\cite{calibration2021}, the most common being temperature scaling~\cite{Guo2017CalibrationOfMNNs}. Finally, the last case is what is being addressed by the recent literature on learning to reject~\cite{Hendrickx2021}, which is currently confined to the academic world, but could greatly benefit from incorporating the notion of value that we introduce here. In this paper, we focus on classifiers that can integrate threshold-based filtering mechanisms, enabling the use of the ``value'' metric with any model capable of providing confidence scores alongside its predictions.



In formalizing ``value'', we will progressively make a few assumptions that i) allow to simplify the presentation of the problem without altering the essence of the concepts, ii) are reasonable in many if not most use cases, and iii) make the definition of the value function easier to understand and interpret for the users who eventually have to deploy ML into their companies. We scope the conversation on classification problems as it makes it easy to ground the examples and terminology, and because it is easier to define a notion of accuracy. This is important: people understand accuracy because it is simple, and that has beneficial value even if accuracy is ``inaccurate'' as a metric, and most users will not be able to express complex value functions. Note however that our results also apply to other performance measures, like F1-score, as we will show in our experimental evaluation.

\subsection{Definition of value}
\label{sec:defValue}
We have a classifier $g$ that operates on test examples $x \in \mathcal{D}$ and returns either a predicted class $y \in \mathcal{Y}$ or a special label $y_r$, denoting ``rejection'' of the prediction. Then, we can compute the average value per the prediction of applying a model $g$ over $\mathcal{D}$ as follows:

\begin{equation}
V(g,\mathcal{D}) = \rho V_r + (1-\rho) (\alpha V_c + (1-\alpha) V_w ) 
\end{equation}
where $\rho$ is the proportion of items in $\mathcal{D}$ that are rejected by $g$ (classified as $y_r$). The term $\alpha$ denotes the accuracy of predictions that exceed the threshold. $V_r$ refers to the value associated with rejecting an item, independently of the correctness of its prediction, and thus resorting to a default path, typically involving a human expert. $V_c$ is the value of correctly classifying an item, which is only granted for non-rejected items. Finally, $V_w$ is the value of an incorrect classification, which again is only granted (or rather, paid) for non-rejected items.  Although these values can be expressed in monetary terms, such as dollars, we focus on their relative values to facilitate comparison between different models and learning strategies.


We define the baseline scenario as one in ML is not utilized, or equivalently, where all predictions are rejected. We set this baseline value to 0 ($V_r = 0$), which simplifies the process of evaluating a model by determining (i) whether it improves upon the baseline, and (ii) whether adopting AI is beneficial for the specific problem at hand.
\begin{equation}
V(g,\mathcal{D}) = (1-\rho) (\alpha V_c + (1-\alpha) V_w ) 
\end{equation}
We also express $V_w$ in terms of $V_c$, as in $V_w = - k V_c$, where $k$ is a constant telling us how bad is an error with respect to getting the correct prediction: 
\begin{equation}
V(g,\mathcal{D}) = V_c (1-\rho) (\alpha  - k(1-\alpha)   ) 
\end{equation}
%
In the value formula, $V_c$ acts as a scaling factor. When evaluating an AI-powered solution workflow, the specific magnitude of this factor is less critical. Instead, we consider the value relative to a unit of $V_c$ dollars, effectively normalizing $V_c$ to focus primarily on value. Thus, we can discuss value in terms of ``value per dollar unit of rejection cost (detrimental value)'' denoted as $V^\prime=V/V_c$. To simplify further without deviating from the equations, we set $V_c = 1$. Therefore, we obtain:
\begin{equation}
\label{eq:simple}
V(g,\mathcal{D}) = (1-\rho) (\alpha - k(1-\alpha) ) 
\end{equation}
%
Equation \ref{eq:simple} embodies the same concepts as Equation 1, streamlining our presentation.
%
%

\subsection{Filtering by threshold} \label{sec:filteringByThreshold}

\begin{figure}[!htbp]
\centering
\includegraphics[width=0.99\textwidth]{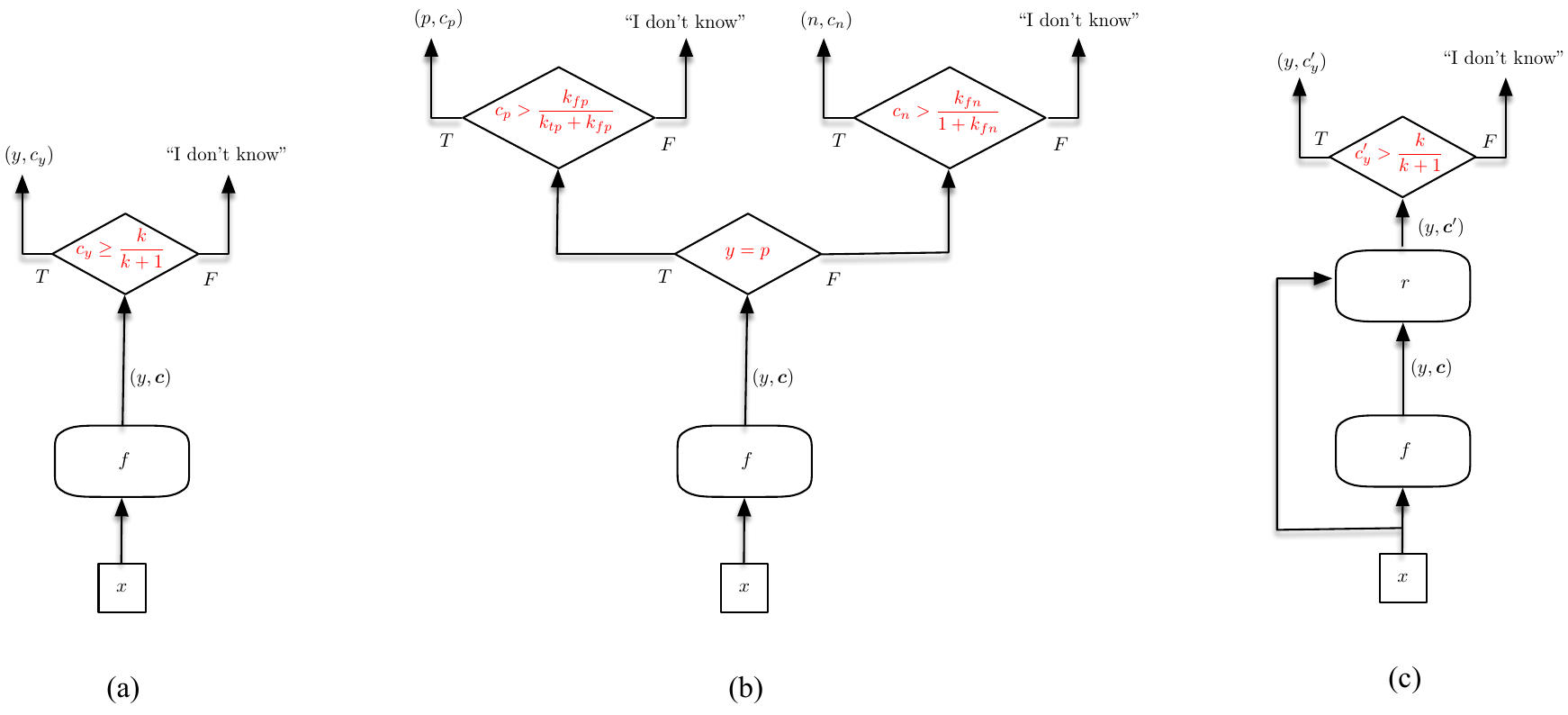}
\caption{Adapting selective classifiers to maximize value: (a) threshold-based selector (b) cost-sensitive threshold-based selector; (c) recalibrator + threshold-based selector. Changes with respect to standard counterparts are highlighted in red.}
\label{fig:value_selector}
\end{figure}

We now focus on the most common situation observed in practice; the model selectivity is applied by thresholding confidence values and rejecting predictions that have confidence $c_y$ less than a threshold $\tau$ (case (a) in Figure~\ref{fig:options}). We are given a model $m$ that processes items $x \in \mathcal{D}$ and returns a vector of confidences (one per class). This is the output of a softmax; for each $x$, we consider the pair $y$, $c_y$ corresponding to the top-level prediction of $m(x)$ and the confidence associated with the prediction. Given a threshold $\tau$, we define a function $s$:
\[   
s(y, c_y, \tau) = 
     \begin{cases}
       \text{$y$} & \text{if} \; \text{$c_y \geq \tau$,}\\
       \text{$y_r$} &\text{otherwise.} \\ 
     \end{cases}
\]
where $y_r$ is the special class label denoting ``rejection'' of the prediction. Our classifier $g$ is therefore now expressed in terms of $m$ and $\tau$. This means that we can express the value as a function of $m,\mathcal{D},\tau$. In a given use case, when we are given $m$ and have knowledge of $k$, we select the threshold $\tau \in [0,1]$ that optimizes $V(g,\mathcal{D})$ (We assume $\tau$ is unique or we randomly pick one if not). Thus, we can express the value of our classification logic as a function of $(m,\mathcal{D},k)$:
\begin{equation}
\label{eq:simplevaluek}
V(m,\mathcal{D},k) =  (1-\rho_\tau) (\alpha_\tau - k (1-\alpha_\tau))  
\end{equation}
Notice that $\tau$ can be set empirically on some tuning dataset $\mathcal{D}$ (it depends on $m,\mathcal{D},k$), and $\rho_\tau$ and $\alpha_\tau$ reflect the proportions $\rho$ and $\alpha$ given $\tau$. 
However, if we are aware of the properties of confidence vectors, we can set $\tau$ regardless of $\mathcal{D}$. For example, if we assume perfect calibration (where the expected accuracy for a prediction of confidence $c$ is $c$)~\cite{calibration2021}, then we know that the threshold is at the point where the value of accepting a prediction is greater than zero, and $\alpha_\tau=\tau$.
This means that to have $V(m,\mathcal{D},k)>0$ we need $\tau - k +k\tau >0 $, which means 
\begin{equation}
\tau > k / (k+1)
\label{eq:theo_thres}
\end{equation}
This conforms to intuition: if k is large, it never makes sense to predict, better go with the default. If k=0 (no cost for errors), we might always predict since there is no penalty for applying inaccurate predictions. Perhaps paradoxically, this case where inaccurate predictions are harmless is when \textit{accuracy} is the metric we want to use. If k=1 (errors are the mirror image of correct predictions), then our threshold is 0.5. 
Figure~\ref{fig:value_selector}(a) shows how a simple threshold-based selector can be adapted to maximize model value. In most real-world settings, especially for complex models, the available classifier will not be perfectly calibrated. In these cases, the threshold can be chosen by either recalibrating the model first using existing recalibration approaches~\cite{calibration2021} and then applying Eq.~(\ref{eq:theo_thres}), or directly maximizing Eq.~(\ref{eq:simplevaluek}) over a separate validation set before testing the classifier. We will evaluate both strategies in our experimental evaluation (see Section~\ref{sec:calibration}).

In deriving the threshold, we initially assumed that all errors incur equal costs. However, we will next demonstrate how this derivation can be readily adapted to cost-sensitive settings.

\subsection{Cost-sensitive value and thresholds}
\label{sec:cost_sensitive}

In this section, we extend the discussion on the value and optimal threshold to the setting in which different errors have different costs (and possibly, different correct predictions have different beneficial values). We focus on the binary classification setting for simplicity, but the reasoning can be easily generalized to multiclass classification. In cost-sensitive learning, the standard approach is that of giving a specific cost to each type of error and correct prediction (in which case the "cost" is the benefit)~\cite{Ling2010}. We adapt this strategy to the value case, by providing a specific value for each possible type of error and correct prediction. The cumulative value of a selective classifier $g$ on a dataset $\mathcal{D}$ can be written as (setting $V_r = 0$ as in the cost-insensitive case):
\begin{align*}
V(g,\mathcal{D}) = (1-\rho) (N_{tp} V_{tp} + N_{tn} V_{tn} + N_{fp} V_{fp} + N_{fn} V_{fn})
\end{align*}
where $N_{tp}, N_{tn}, N_{fp}, N_{fn}$ are the numbers of true positives, true negatives, false positives, and false negatives in $\mathcal{D}$, and 
$V_{tp}, V_{tn}, V_{fp}, V_{fn}$ are the values associated to the corresponding predictions. Let $V_{c}$ be the base cost for a correct prediction. This is typically associated with a correctly predicted negative instance, i.e., $V_{tn} = V_{c}$. We can define the other values as multiples of this base cost as follows:
\begin{displaymath}
    V_{tp} =  k_{tp} V_{c}, \quad  
    V_{fp} =  - k_{fp} V_{c}, \quad
    V_{fn} =  - k_{fn} V_{c}
\end{displaymath}
\noindent for some user-defined and application-specific constants $k_{tp}, k_{fp}, k_{fn}$. The cumulative value simplifies as:
\begin{align*}
V(g,\mathcal{D}) & = (1-\rho) (N_{tp} k_{tp} V_{c} + N_{tn} V_{c} - N_{fp} k_{fp} V_{c} - N_{fn} k_{fn} V_{c}) \\
& = (1-\rho) V_{c} (k_{tp} N_{tp} + N_{tn} - k_{fp} N_{fp}  - k_{fn} N_{fn}) 
\end{align*}
Setting $V_{c} = 1$ (unit of value) as in the cost-insensitive case, we get:
\begin{align*}
V(g,\mathcal{D}) & = (1-\rho) (k_{tp} N_{tp} + N_{tn} - k_{fp} N_{fp} - k_{fn} N_{fn}) 
\end{align*}
Let's now focus on the standard setting of a classifier rejecting by threshold. Note that we need to set class-specific thresholds $\tau_{p}$ and $\tau_{n}$ for positive and negative predictions respectively to account for the different costs. Consider an instance $x$ predicted as positive by the classifier. Its expected value (according to the predictions in $\mathcal{D}$) is given by:
\begin{align*}
V(g,x) & = (1-\rho) (k_{tp} N_{tp} / N_{p} - k_{fp} (N_{fp} / N_{p}) \\
& = (1-\rho) (k_{tp} N_{tp} / N_{p} - k_{fp} (1 - N_{tp}/N_{p})) \\
& = (1-\rho) (N_{tp} / N_{p} (k_{tp} + k_{fp}) - k_{fp})
\end{align*}
\noindent where we normalized $N_{tp}$ and $N_{fp}$ by $N_{p}$, the number of positive instances in $\mathcal{D}$, to turn them into probabilities, and we removed the terms containing $N_{tn}$ and $N_{fn}$ as their corresponding probabilities are zero if the instance is predicted as positive. 

If the classifier is perfectly calibrated, we know that $N_{tp} / N_{p} = \tau_{p}$. A positive value for the instance is thus achieved by setting $\tau_{p}$ as:
\begin{equation}
    \tau_{p} > \frac{k_{fp}}{k_{tp} + k_{fp}}
    \label{eq:theo_thres_pos}
\end{equation}
Similarly, if $x$ is predicted as negative by the classifier, it is expected value is given by:
\begin{align*}
V(g,x) & = (1-\rho) (N_{tn} / N_{n} - k_{fn} N_{fn} / N_{n}) \\
& = (1-\rho) (N_{tn} / N_{n} - k_{fn} (1 - N_{tn} / N_{n})) \\
& = (1-\rho) (N_{tn} / N_{n} (1 + k_{fn}) - k_{fn})
\end{align*}
\noindent where $N_{n}$ is the number of negative instances in the training set. If the classifier is perfectly calibrated, we know that $N_{tn} / N_{n} = \tau_{n}$. A positive value for the instance is thus achieved by setting $\tau_{n}$ as:
\begin{equation}
    \tau_{n} > \frac{k_{fn}}{1 + k_{fn}}
    \label{eq:theo_thres_neg}
\end{equation}

Figure~\ref{fig:value_selector}(b) shows how to adjust a threshold-based selector to maximize value in a cost-sensitive setting. We assumed a binary classification setting for simplicity, but the derivation can be easily extended to account for class-specific thresholds in multiclass classification.

\section{Experiments} \label{sec:expWork}

We now explore how adopting a value-oriented perspective influences model evaluation and application. Specifically, we aim to address the following questions:
\begin{itemize}
    \item[{\bf Q1}] Is model accuracy (or F1-score) a sensible indicator of the value of a model?
    
    \item[{\bf Q2}] Is cost-sensitive error a sensible indicator of the value of a model in cost-sensitive settings ? 

    \item[{\bf Q3}] How does calibration affect the value of a model?

    \item[{\bf Q4}] How does predicting in an out-of-distribution setting affect the value of a model?
\end{itemize}

Our experimental evaluation is focused on NLP classification tasks, for which we analyze the behavior of simple as well as state-of-the-art models over various datasets, models, and text encoders. This choice
stems from the broad diffusion of NLP models in companies, and from our experience (see Section \ref{insights}) in industrial use cases that were all NLP-based. However, the concept of value can be applied to any ML model deployed in a practical application, and we believe that the main results of our experimental evaluation hold for many other domains. We refer the reader to our GitHub repo\footnote{\href{https://github.com/burcusayin/value-of-ml-models/}{https://github.com/burcusayin/value-of-ml-models/}} for the companion code. 

\subsection{Experimental Setup}
\label{subsec:ExpSetup}

\begin{table*}[ht]
    \centering
    \scalebox{0.83}{%
    \begin{tabular}{llll}

    \textbf{Task} &  \textbf{Dataset} & \textbf{Train/Val/Test size} \\
    \hline
    Classifying tweets as ``hate'', and ``non-hate'' (binary)  & Hate Speech & 7734/967/967 \\

    Classify Twitter posts to detect clickbait (binary) & Clickbait &  17600/4395/18979 \\

    Sentiment analysis on Amazon product reviews & MDS Electronics & 2000/200/3386 \\
    (3-class; positive, negative, and neutral) &   MDS DVD & 2000/200/4265 \\
     &   MDS Books & 2000/200/5481 \\
     &   MDS Kitchen & 2000/200/5745 \\   
    \hline
    \end{tabular}%
    }
    \caption{Statistics of the datasets used in the experiments}
    \label{tab:datasets}
\end{table*}

\noindent\textbf{Datasets and Tasks}  Table~\ref{tab:datasets} presents a summary of the characteristics of the datasets we employed and their corresponding classification tasks. Additional information is provided in the following.

\begin{itemize}
    \item \emph{Hate-speech detection on Twitter.} 
    We replicated the original tests from \cite{Arango2019HateSpeech} where we analyzed two widely used models (\cite{Agrawal2018HS,Badjatiya2017HS}) and tested them on the Waseem et al. \cite{Waseem2016HatefulSO} dataset. However, we could only recover 9668 of the tweets as of October 2021 (the dataset size is 14949 in the original paper). 
  
    \item \emph{Clickbait detection.} The \textit{Clickbait Challenge} on the \textit{Webis Clickbait Corpus 2017}\footnote{\href{https://webis.de/data/webis-clickbait-17.html}{https://webis.de/data/webis-clickbait-17.html}} was classifying Twitter posts as a clickbait or not. Both training and test sets are publicly available\footnote{\href{https://zenodo.org/record/5530410\#.YWcFtC8RrRV}{https://zenodo.org/record/5530410\#.YWcFtC8RrRV}}, while each team was free to choose a subset of the training set for validation (we followed the ``blobfish'' team).
    
    \item \emph{Multi-Domain Sentiment Analysis - and Dataset (MDS).} Sentiment analysis based on a dataset for domain adaptation.\footnote{\href{http://nlpprogress.com/english/domain\_adaptation.html}{http://nlpprogress.com/english/domain\_adaptation.html}} The data includes four categories of Amazon products (DVD, Books, Electronics, and Kitchen). The task is to learn sentiment from one of these domains and test it on the others. 
\end{itemize}

\noindent\textbf{Models and text encoders.} For each task in our experiments, we use different models (see Table \ref{tab:expOverview} and the accompanying code repository for details). Since we do not train models and use the validation set only to determine the optimal threshold, we do not perform standard cross-validation. The optimal threshold is selected by evaluating the model on the validation set across a range of candidate thresholds and choosing the one that maximizes performance. 


\begin{itemize}
\item For the hate-speech dataset, we test the following leaderboard
  models: (i) {\em Badjatiya et al.} \cite{Badjatiya2017HS} which uses an RNN
  to construct word embeddings
  and then classify them with Gradient-Boosted Decision Tree. In the
  original paper, test accuracy is measured as the average of the ten
  folds in cross-validation; however, in our reproduction, we
  separated validation and test set before cross-validation, and they
  are used for evaluation only after training. (ii) one model from
  {\em Agrawal and Awekar} \cite{Agrawal2018HS} which is composed of an
  embedding layer 
  followed by a Bidirectional LSTM and a fully connected layer with softmax activation.

\item For the clickbait detection dataset, we test 4 models from one
  leaderboard team on clickbait challenge: {\em fullnetconc}, {\em weNet},
  {\em lingNet}, and {\em fullNet} which are published on
  Github.\footnote{\href{https://github.com/clickbait-challenge/blobfish}{https://github.com/clickbait-challenge/blobfish}} This team
  modified the task into binary classification - they categorized
  items with a score under 0.5 into ``non-clickbaiting", and vice
  versa.

\item For the MDS dataset, we referred to the leaderboard for the
  sentiment analysis task of Domain
  adaptation\footnote{\href{http://nlpprogress.com/english/domain_adaptation.html}{nlpprogress.com/english/domain\_adaptation.html}}
  and tested the best-performing leader-board model,
  {\em Multi-task tri-training (mttri)} by Ruder
  et. al. \cite{Ruder2018}, that leverages multi-task learning strategies 
  to improve the performance of tri-training. As the source code of other 
  competing approaches was not publicly available, we compared mttri with three baseline
  models from the scikit-learn library\footnote{\href{https://scikit-learn.org/}{https://scikit-learn.org/}}: (i) a simple Logistic Regression model ({\em LogR}); (ii) a basic MLP with a single hidden layer ({\em MLP1}); (iii) an MLP with four hidden layers ({\em MLP4}).
  All models where tested with a simple \textit{TF-IDF} encoding.
\end{itemize}

\begin{table*}[!htbp]
\centering
\scalebox{0.85}{%
\begin{tabular}{lll}

\textbf{Dataset} & \textbf{Models} & \textbf{Model details}
\\
\hline
    Hate-speech detection & \cite{Badjatiya2017HS}, \cite{Agrawal2018HS} & Leader-board models\\
    Clickbait detection & fullnetconc, weNet, lingNet, fullNet & Leader-board models\\ 
    
    MDS & mttri \cite{Ruder2018}& Leader-board\\
    & Google's T5-base & fine-tuned for sentiment analysis \\
    & SieBERT & fine-tuned RoBERTa-large \\
    & LogR, MLP1, MLP4 & from scikit-learn library \\
    & GPT-3 & fine-tuned for sentiment analysis \\
    \hline
    \end{tabular}%
    }
    \caption{Models used in the experiments}
    \label{tab:expOverview}
\end{table*}

\subsection{Results}

\begin{table}[!b]
  \begin{center}
    \begin{sc}
    \scalebox{0.85}{%
    \begin{tabular}{llllllllll}
    \hline
    \multirow{2}{*}{\textbf{Task}} &
    \multirow{2}{*}{\textbf{Model}} & \multirow{2}{*}{\textbf{Accuracy}} & \multirow{2}{*}{\textbf{F1}}&  \multicolumn{6}{c}{\textbf{Value}}  \\
    & & & & $k=0$ & $k=1$ & $k=2$ & $k=4$ & $k=8$ & $k=10$ \\ 
       \hline
        Hate Speech  & Badj. et al & \textbf{0.822} & \textbf{0.626} & \textbf{0.822} & \textbf{0.644} & \textbf{0.51} & \textbf{0.362} & \textbf{0.272} & \textbf{0.217} \\
        & Agr. et al. & 0.732 & 0.621 &  0.732 & 0.464 & 0.22 & -0.213 & -1.081 & -1.499\\ 
       \hline
       Clickbait & fullnetconc & \textbf{0.857} & \textbf{0.684} & \textbf{0.857} & \textbf{0.715} & 0.564 & 0.286 & 0.041 & 0.013 \\ 
       & weNet & 0.852 & 0.672 & 0.852 & 0.703 & 0.561 & 0.306 & 0.04 & 0.011 \\ 
       & lingNet & 0.82 & 0.565 & 0.82 & 0.64 & 0.442 & 0.079 & 0.0 & 0.0 \\ 
       & fullNet & 0.856 & 0.663 & 0.856 & 0.713 & \textbf{0.588} & \textbf{0.367} & \textbf{0.061} & \textbf{0.015} \\
      \hline
      MDS Electronics & LogReg & 0.762 & 0.736 & 0.762 & 0.524 & 0.339 & 0.162 & 0.053 & 0.033 \\
      & MLP1 & 0.749 & 0.711 & 0.749 & 0.497 & 0.327 & \textbf{0.18} & \textbf{0.081} & \textbf{0.062} \\
      & MLP4 & 0.735 & 0.713 & 0.735 & 0.47 & 0.24 & -0.143 & -0.78 & -1.06 \\
      & mttri & \textbf{0.808} & \textbf{0.786} & \textbf{0.808} & \textbf{0.616} & \textbf{0.441} & 0.148 & -0.354 & -0.58 \\
        \hline
      MDS DVD & LogReg & 0.74 & \textbf{0.739} & 0.74 & 0.48 & \textbf{0.283} & 0.122 & 0.038 & 0.027 \\
      & MLP1 & 0.728 & 0.732 & 0.728 & 0.457 & 0.274 & \textbf{0.133} & \textbf{0.054} & \textbf{0.038} \\
      & MLP4 & 0.72 & 0.724 & 0.72 & 0.439 & 0.202 & -0.158 & -0.737 & -0.981 \\
      & mttri & \textbf{0.753} & 0.725 & \textbf{0.753} & \textbf{0.506} & 0.28 & -0.123 & -0.84 & -1.166 \\
      \hline
      MDS Books & LogReg & 0.704 & 0.678 & 0.704 & 0.408 & 0.228 & \textbf{0.102} & \textbf{0.022} & \textbf{0.015} \\
      & MLP1 & 0.691 & 0.662 & 0.691 & 0.382 & 0.134 & 0.013 & -0.017 & -0.013 \\
      & MLP4 & 0.696 & 0.681 & 0.696 & 0.393 & 0.154 & -0.171 & -0.666 & -0.86 \\
      & mttri & \textbf{0.742} & \textbf{0.712} & \textbf{0.742} & \textbf{0.484} & \textbf{0.254} & -0.16 & -0.869 & -1.215 \\
      \hline 
      MDS Kitchen & LogReg & 0.782 & 0.771 & 0.782 & 0.565 & 0.374 & 0.176 & 0.06 & 0.034 \\
      & MLP1 &  0.765 & 0.752 &  0.765 & 0.53 & 0.337 & 0.164 & \textbf{0.07} & \textbf{0.044} \\
      & MLP4 & 0.761 & 0.758 & 0.761 & 0.521 & 0.312 & 0.003 & -0.478 & -0.685 \\
      & mttri & \textbf{0.821} & \textbf{0.832} & \textbf{0.821} & \textbf{0.642} & \textbf{0.489} & \textbf{0.235} & -0.192 & -0.384 \\
      \hline
    \end{tabular}%
    }
    \caption{Accuracy and F1-score results compared with value computed for increasing values of the cost factor $k$. For each dataset and metric, the best performance is highlighted in bold.}
    \label{tab:q1}
    \end{sc}
  \end{center}
\end{table}

\subsubsection{Q1: Accuracy and F1-score are poor indicators of model value}

We first investigate whether standard performance metrics, like accuracy and F1-score, are sensible indicators of the value of the model, and how this depends on the magnitude of the cost factor $k$. Following the simplification in Section~\ref{sec:defValue}, we set $V_r = 0$ and $V_c = 1$, and use the threshold in Eq.~\ref{eq:theo_thres} to decide whether to accept or reject each prediction given a certain $k$. 

Table~\ref{tab:q1} report results in terms of accuracy, F1-score, and value for different values of $k \in [0,10]$. As expected, the value of a model decreases substantially with the increase of the cost factor, with many models achieving {\em negative} value for larger values of $k$. Note that a model is useful only if its value exceeds 0; otherwise, it is deemed unnecessary, and the system can proceed without it. We want to stress that the cost factors we considered are fairly small and definitely realistic. For instance, setting  $k=4$ means that ``being wrong is 4 times as bad'' with respect to the advantage of being right. Many scenarios have values of $k$ way more extreme (e.g., in medical decision support systems~\cite{sutton2020}). Notice that accuracy corresponds to the case where we do not reject any predictions, which corresponds to setting $k=0$, a rather unrealistic scenario. 

Another major finding is that accuracy is a quite poor proxy of value even in relative terms. Boldface numbers indicate the best performing model in terms of the different metrics. It is clear that the best performing model is largely dependent on the cost factor, and that accuracy quickly becomes totally unreliable as a metric to identify the most appropriate model to employ. Replacing accuracy with F1-score does not change much. While we do observe substantially lower values for the unbalanced datasets (Hate Speech and Clickbait), the best performing model is unchanged almost everywhere. 

\begin{table}[!b]
  \begin{center}
    \begin{sc}
    \scalebox{0.755}{%
    \begin{tabular}{llllllllllll}
    \hline
    \multirow{2}{*}{\textbf{Task}} &
    \multirow{2}{*}{\textbf{Model}} & \multicolumn{5}{c}{\textbf{Cost-sensitive error}} &  \multicolumn{5}{c}{\textbf{Value}}  \\
    & & $k=1$ & $k=2$ & $k=4$ & $k=8$ & $k=10$ & $k=1$ & $k=2$ & $k=4$ & $k=8$ & $k=10$ \\ 
       \hline  
        Hate Speech & Badj. et al & \textbf{0.178} & \textbf{0.297} &  0.535 &  1.01 &  1.248 & \textbf{0.644} & \textbf{0.545} & \textbf{0.389} & \textbf{0.315} & \textbf{0.278} \\
        & Agr. et al. &  0.268 &  0.322 &  \textbf{0.429} &  \textbf{0.644} &  \textbf{0.752} & 0.464 & 0.405 & 0.32 & 0.157 & 0.098 \\ 
       \hline
       Clickbait & fullnetconc & \textbf{0.143} &  \textbf{0.221} &  \textbf{0.377} &  \textbf{0.689} &  \textbf{0.845} & \textbf{0.715} & 0.608 & 0.368 & 0.131 & \textbf{0.103} \\
    	& weNet & 0.148 &  0.228 &  0.388 &  0.707 &  0.867 &  0.703 & 0.604 & 0.381 & 0.124 & 0.094 \\
    	& lingNet & 0.18 &  0.295 &  0.524 &  0.983 &  1.213 & 0.64 & 0.467 & 0.125 & 0.052 & 0.052  \\
    	& fullNet & 0.144 &  0.234 &  0.416 &  0.779 &  0.961 & 0.713 & \textbf{0.631} & \textbf{0.446} & \textbf{0.15} & \textbf{0.103}\\
  
      \hline
      MDS Electronics & LogReg & 0.238 & 0.406 & 0.742 & 1.413 & 1.749 & 0.524 & 0.442 & \textbf{0.355} & \textbf{0.293} & \textbf{0.282} \\
  	& MLP1 & 0.259 & 0.436 & 0.791 & 1.5 & 1.854 & 0.497 & 0.413 & 0.338 & 0.284 & 0.274 \\
  	& MLP4 & 0.254 & 0.418 & 0.745 & 1.4 & 1.727 & 0.47 & 0.33 & 0.09 & -0.313 & -0.492 \\
  	& mttri & \textbf{0.192} & \textbf{0.33} & \textbf{0.607} & \textbf{1.159} & \textbf{1.436} & \textbf{0.616} & \textbf{0.495} & 0.286 & -0.085 & -0.245 \\
        \hline
      MDS DVD & LogReg & 0.26 & 0.394 & 0.663 & 1.201 & 1.47 & 0.48 & \textbf{0.375} & 0.295 & 0.255 & \textbf{0.251} \\
  	& MLP1 & 0.271 & 0.404 & 0.67 & 1.203 & 1.469 &  0.457 & 0.36 & \textbf{0.298} & \textbf{0.26} & \textbf{0.251}  \\
  	& MLP4 & 0.278 & \textbf{0.392} & \textbf{0.62} & \textbf{1.075} & \textbf{1.303} &  0.439 & 0.327 & 0.16 & -0.089 & -0.193 \\
  	& mttri & \textbf{0.247} & 0.412 & 0.744 & 1.406 & 1.737 & \textbf{0.506} & 0.352 & 0.072 & -0.431 & -0.663  \\
      \hline
      MDS Books & LogReg & 0.296 & 0.489 & 0.874 & 1.645 & 2.03 &  0.408 & \textbf{0.332} & \textbf{0.269} & \textbf{0.222} & \textbf{0.219} \\
  	& MLP1 & 0.303 & 0.492 & 0.87 & 1.627 & 2.005 & 0.382 & 0.272 & 0.197 & 0.18 & 0.183 \\
  	& MLP4 & 0.312 & 0.486 & \textbf{0.832} & \textbf{1.525} & \textbf{1.871} &  0.393 & 0.258 & 0.081 & -0.183 & -0.283 \\
  	& mttri & \textbf{0.258} & \textbf{0.45} & 0.834 & 1.603 & 1.987 &  \textbf{0.484} & 0.32 & 0.018 & -0.52 & -0.789 \\
      \hline 
      MDS Kitchen & LogReg & 0.218 & 0.345 & 0.599 & 1.108 & 1.363 & 0.565 & 0.466 & 0.365 & 0.306 & 0.295 \\
  	& MLP1 & 0.242 & 0.375 & 0.64 & 1.171 & 1.436 & 0.53 & 0.433 & 0.339 & 0.292 & 0.279  \\
  	& MLP4 & 0.248 & 0.387 & 0.665 & 1.22 & 1.498 & 0.521 & 0.416 & 0.263 & 0.026 & -0.076 \\
  	& mttri & \textbf{0.179} & \textbf{0.238} & \textbf{0.355} & \textbf{0.59} & \textbf{0.708} & \textbf{0.642} & \textbf{0.589} & \textbf{0.503} & \textbf{0.376} & \textbf{0.31}  \\
      \hline
    \end{tabular}%
    }
    \caption{Comparison between cost-sensitive error and value for different values of $k=k_{fn}$ (with $k_{fp}=1$). For each dataset and metric, the best performance is highlighted in bold.}
    \label{tab:costsensitive}
    \end{sc}
  \end{center}
\end{table}

\subsubsection{Q2: Cost-sensitive error is a poor indicator of model value in cost-sensitive settings}
\label{sec:costsensitive_res}
The previous evaluation assumed equal cost for the different types of error. This is however rarely the case in practical applications, where false negative errors (e.g., undiagnosed diseases) can be far more costly than false positive ones (i.e., false alarms). Section~\ref{sec:cost_sensitive} shows how to adapt value to this cost-sensitive setting, and how to determine cost-sensitive thresholds that are specific for each predicted class. In the following we evaluate the value of models in this cost-sensitive setting. We replace accuracy and F1, which are clearly inappropriate in this setting, with cost-sensitive error~\cite{elkan2001} a popular performance measure in the cost-sensitive learning literature. Cost-sensitive error is obtained by computing the weighted sum of errors, with the weights given by the corresponding cost, i.e. $(N_{fn}k_{fn} + N_{fp}k_{fp})/|\mathcal{D}|$, where we divide by $|\mathcal{D}|$ to remove the dependency on the size of the dataset. For simplicity, and consistently with common practice in the literature, we set $k_{fp}=1$ and vary $k_{fn} \in [1,10]$. Results are shown in Table~\ref{tab:costsensitive}. While cost-sensitive error identifies different best performing models for different values of the cost, in only one case (MDS Kitchen) it consistently agrees with value across the spectrum of costs.
What is worse, for large values of $k_{fn}$ it often detects as best performing models that actually achieve {\em negative} value, making it a poor overall indicator of model value. The problem is not how it treats the costs of different errors, but in the fact that it does not assume a selective classifier and a corresponding cost-sensitive rejection threshold, which is the main practical contribution of our definition of value. This also implies that cost sensitive learning~\cite{He2013}, that aims at training classifiers to minimize (a certain notion of) cost-sensitive error, should be coupled with learning to reject mechanisms~\cite{Hendrickx2021} in order to be fully effective in optimizing the value of the learned models.

\begin{table}[!b]
  \begin{center}
    \begin{sc}
    \scalebox{0.85}{%
    \begin{tabular}{llllllllll}
    \hline
    \multirow{2}{*}{\textbf{Task}} &
    \multirow{2}{*}{\textbf{Model}} & \multirow{2}{*}{\textbf{Accuracy}} & \multirow{2}{*}{\textbf{F1}}&  \multicolumn{6}{c}{\textbf{Value}}  \\
    & & & & $k=0$ &$k=1$ & $k=2$ & $k=4$ & $k=8$ & $k=10$ \\ 
       \hline  
        Hate Speech & Badj. et al & \textbf{0.822} & \textbf{0.626} & \textbf{0.822} & \textbf{0.644} & \textbf{0.513} & \textbf{0.359} & \textbf{0.268} & \textbf{0.218} \\
        & Agr. et al. &  0.732 & 0.621 &  0.732 & 0.464 & 0.207 & 0.0 & 0.0 & 0.0\\ 
       \hline
       Clickbait & fullnetconc & \textbf{0.857} & \textbf{0.684} & \textbf{0.857} & \textbf{0.715} & \textbf{0.608} & \textbf{0.488} & \textbf{0.374} & 0.331\\
    	& weNet & 0.852 & 0.672 & 0.852 & 0.703 & 0.597 & 0.472 & 0.357 & 0.326  \\
    	& lingNet & 0.82 & 0.565 & 0.82 & 0.64 & 0.499 & 0.348 & 0.173 & 0.115  \\
    	& fullNet & 0.856 & 0.663 & 0.856 & 0.713 & 0.6 & \textbf{0.488} & 0.372 & \textbf{0.335}\\
  
      \hline
      MDS Electronics & LogReg & 0.762 & 0.736 & 0.762 & 0.524 & 0.362 & \textbf{0.226} & \textbf{0.119} & \textbf{0.098}\\
  	& MLP1 &  0.745 & 0.711 &  0.745 & 0.491 & 0.33 & 0.174 & 0.096 & 0.062 \\
  	& MLP4 & 0.745 & 0.713 & 0.745 & 0.491 & 0.291 & 0.11 & 0.0 & 0.0 \\
  	& mttri & \textbf{0.808} & \textbf{0.786} & \textbf{0.808} & \textbf{0.616} & \textbf{0.447} & 0.192 & 0.112 & 0.0 \\
        \hline
      MDS DVD & LogReg & 0.74 & \textbf{0.739} & 0.74 & 0.48 & \textbf{0.315} & \textbf{0.17} & \textbf{0.09} & \textbf{0.062} \\
  	& MLP1 & 0.729 & 0.732 & 0.729 & 0.459 & 0.28 & 0.148 & 0.037 & 0.023  \\
  	& MLP4 & 0.722 & 0.724 & 0.722 & 0.443 & 0.235 & 0.056 & 0.0 & 0.0  \\
  	& mttri & \textbf{0.753} & 0.725 & \textbf{0.753} & \textbf{0.506} & 0.292 & 0.08 & 0.0 & 0.0  \\
      \hline
      MDS Books & LogReg & 0.704 & 0.678 & 0.704 & 0.408 & 0.234 & \textbf{0.111} & \textbf{0.01} & \textbf{0.001}  \\
  	& MLP1 & 0.697 & 0.662 & 0.697 & 0.395 & 0.199 & 0.002 & 0.0 & 0.0 \\
  	& MLP4 & 0.688 & 0.681 & 0.688 &  0.375 & 0.095 & 0.0 & 0.0 & 0.0  \\
  	& mttri & \textbf{0.742} & \textbf{0.712} & \textbf{0.742} & \textbf{0.484} & \textbf{0.264} & -0.011 & 0.0 & 0.0 \\
      \hline 
      MDS Kitchen & LogReg & 0.782 & 0.771 & 0.782 & 0.565 & 0.41 & \textbf{0.267} & \textbf{0.153} & \textbf{0.127} \\
  	& MLP1 & 0.758 & 0.752 & 0.758 & 0.515 & 0.345 & 0.197 & 0.096 & 0.011  \\
  	& MLP4 & 0.752 & 0.758 & 0.752 & 0.504 & 0.305 & 0.122 & 0.0 & 0.0 \\
  	& mttri & \textbf{0.821} & \textbf{0.832} & \textbf{0.821} & \textbf{0.642} & \textbf{0.493} & 0.227 & 0.102 & 0.0  \\
      \hline
    \end{tabular}%
    }
    \caption{Comparison between accuracy, F1-score and value for recalibrated models. For each dataset and metric, the best performance is highlighted in bold.}  
    \label{tab:calibration}
    \end{sc}
  \end{center}
\end{table}

\subsubsection{Q3: Lack of calibration substantially affects model value}
\label{sec:calibration}

The threshold in Eq.~\ref{eq:theo_thres} assumes that models are perfectly calibrated, which is often far from being true for trained models, and deep learning models in particular~\cite{Guo2017CalibrationOfMNNs}. In order to evaluate the role of calibration in determining value of a model, we apply temperature scaling~\cite{Guo2017CalibrationOfMNNs}, a simple yet effective recalibration technique, to each model before applying the threshold (the resulting selector is shown in Figure~\ref{fig:value_selector}(c)). Table~\ref{tab:calibration} reports the results in exactly the same setting as Table~\ref{tab:q1}, but using recalibrated models. Notice that accuracy and F1-score are unchanged, as temperature scaling affects the confidence in the prediction but not how classes are being ranked. In terms of value, however, we observe an overall improvement, quite substantial for larger values of $k$. Note that the effectiveness of calibration in improving the model's ``value'' depends on the accuracy of the calibrated model. The degenerate behaviour of models with negative values is almost completely eliminated, with ``useless" models receiving a value of zero, as expected. These results suggest that learning models should always be recalibrated before being incorporated in practical workflows. This does not mean that one can then resort on standard accuracy or F1-score to choose which model to employ. The best performing model is still largely dependent on the cost factor. Notice that in the domain adaptation scenarios (MDS tasks), simple logistic regression (LogReg) consistently outperforms all other models for large values of $k$. This result should not be unexpected. Logistic regression is known to be a well-calibrated model per-se~\cite{Kull2017BeyondSH}, and temperature scaling likely further improves this behaviour, while more complex models struggle to achieve comparable calibration with simple recalibration strategies. The lively research area of calibration in machine learning and especially deep learning can provide useful solutions to this problem~\cite{calibration2021}.

\subsubsection{Q4: Operating in an out-of-distribution setting substantially affects model value} \label{sec:ood}

The lack of calibration in machine learning models is known to be particularly harmful when the model operates in an out-of-distribution (OOD) setting~\cite{Tomani2019TowardsTP,Wu2022Distribution}, and the results on the domain adaptation tasks in Table~\ref{tab:calibration} confirm this issue. To better understand the role of the OOD setting in determining the value of models, we thus focused on the MDS tasks and complemented the set of models presented in Table~\ref{tab:calibration} with some state-of-the-art transformer models, which should be less affected by the problem given the huge corpora on which they are trained. The transformer models that we employed are the following:

\begin{itemize}
\item Google's T5-base\footnote{\href{https://tinyurl.com/t5-base-finetuned-sentiment}{https://tinyurl.com/t5-base-finetuned-sentiment}}
  \cite{Raffel2020T5} (12-layers, 768-hidden-state, 3072 feed-forward
  hidden-state, 12-heads, 220M parameters) fine-tuned on IMDB
  dataset\footnote{\href{https://huggingface.co/datasets/stanfordnlp/imdb}{https://huggingface.co/datasets/stanfordnlp/imdb}}
  \cite{maas2011Learning} for sentiment analysis task.
\item SieBERT\footnote{\href{https://tinyurl.com/SieBERT-sentiment}{https://tinyurl.com/SieBERT-sentiment}}
\cite{heitmann2020}: a fine-tuned version of
RoBERTa-large\footnote{\href{https://huggingface.co/FacebookAI/roberta-large}{https://huggingface.co/FacebookAI/roberta-large}} model
\cite{liu2021Roberta} (24-layer, 1024-hidden-state, 16-heads, 355M
parameters) for sentiment analysis task that is fine-tuned and
evaluated on 15 diverse text sources. 
\item  GPT-3~\cite{GPT3}. Since it is producing human-like
text for a given input, we fine-tuned it using the OpenAI
API\footnote{\href{https://openai.com/api/}{https://openai.com/api/}}. First, we prepared the MDS
dataset for GPT-3; we cleaned sentences that have more than 2049
tokens, and renamed the text column as ``\textit{prompt}" and the
ground truth column as ``\textit{completion}". Then, we used OpenAI API
to fine-tune GPT-3 separately on each of the 4 domains (DVD, books,
electronics, and kitchen). We specified ``$classification\_n\_classes$"
parameter as 2 and $classification\_positive\_class$ as `1' so that
the API tunes GPT-3 for binary sentiment analysis. Fine-tuning 4
models on the MDS dataset costs a total of \$7.15. In order to test
the fine-tuned models on different target domains, we specified the
\textit{prompt} in the format of ``sentence + -> " because the API
itself uses `` ->" sign to teach GPT-3 that the sentiment for a
\textit{prompt} is (` ->') the \textit{completion}. Thus, fine-tuned
GPT-3 models produce either 0 or 1 for the given input. Testing each
fine-tuned model on the other 3 domains (so, 12 cases in total) costs
\$43.89. We provide our source code on
Github\footnote{\href{https://github.com/burcusayin/value-of-ml-models/}{https://github.com/burcusayin/value-of-ml-models/}} to
show every step of using GPT-3 in our experiments.
\end{itemize}

\begin{table}[!h]
  \begin{center}
    \begin{sc}
    \scalebox{0.9}{%
    \begin{tabular}{lllllllll}
    \hline
    \multirow{2}{*}{\textbf{Task}} &
    \multirow{2}{*}{\textbf{Model}} & \multirow{2}{*}{\textbf{Accuracy}} & \multirow{2}{*}{\textbf{F1}}&  \multicolumn{5}{c}{\textbf{Value}}  \\
    & & & & $k=1$ & $k=2$ & $k=4$ & $k=8$ & $k=10$ \\ 
       \hline
      MDS Electronics & LogReg & 0.762 & 0.736 & 0.524 & 0.339 & 0.162 & 0.053 & 0.033 \\
      & MLP1 & 0.745 & 0.711 & 0.497 & 0.327 & 0.18 & 0.081 & \textbf{0.062} \\
      & MLP4 & 0.745 & 0.713 & 0.47 & 0.24 & -0.143 & -0.78 & -1.06 \\
      & mttri & 0.808 & 0.786 & 0.616 & 0.441 & 0.148 & -0.354 & -0.58 \\
      & T5 & 0.784 & 0.765 & 0.568 & 0.352 & -0.08 & -0.944 & -1.376 \\
  	& SieBERT & \textbf{0.842} & \textbf{0.831} & \textbf{0.685} & \textbf{0.527} & 0.217 & -0.397 & -0.705 \\
  	& GPT-3 & 0.82 & 0.803 & 0.641 & 0.499 & \textbf{0.322} & \textbf{0.127} & 0.051  \\
        \hline
      MDS DVD & LogReg & 0.74 & 0.739 & 0.48 & 0.283 & 0.122 & 0.038 & 0.027 \\
      & MLP1 & 0.729 & 0.732 & 0.457 & 0.274 & 0.133 & 0.054 & 0.038 \\
      & MLP4 & 0.722 & 0.724 & 0.439 & 0.202 & -0.158 & -0.737 & -0.981 \\
      & mttri & 0.753 & 0.725 & 0.506 & 0.28 & -0.123 & -0.84 & -1.166 \\
      & T5 & 0.789 & 0.788 & 0.578 & 0.367 & -0.056 & -0.9 & -1.323 \\
  	& SieBERT & \textbf{0.836} & \textbf{0.832} & \textbf{0.672} & 0.508 & 0.193 & -0.436 & -0.747 \\
  	& GPT-3 & 0.832 & 0.825 & 0.664 & \textbf{0.534} & \textbf{0.367} & \textbf{0.164} & \textbf{0.089}  \\
      \hline
      MDS Books & LogReg & 0.704 & 0.678 & 0.408 & 0.228 & 0.102 & 0.022 & \textbf{0.015} \\
      & MLP1 & 0.697 & 0.662 & 0.382 & 0.134 & 0.013 & -0.017 & -0.013 \\
      & MLP4 & 0.688 & 0.681 & 0.393 & 0.154 & -0.171 & -0.666 & -0.86 \\
      & mttri & 0.742 & 0.712 & 0.484 & 0.254 & -0.16 & -0.869 & -1.215 \\
      & T5 & 0.77 & 0.791 & 0.541 & 0.311 & -0.148 & -1.066 & -1.525 \\
  	& SieBERT & \textbf{0.826} & \textbf{0.827} & \textbf{0.652} & \textbf{0.479} & 0.136 & -0.547 & -0.879  \\
  	& GPT-3 & 0.806 & 0.808 & 0.613 & 0.46 & \textbf{0.272} & \textbf{0.077} & 0.004  \\
      \hline 
      MDS Kitchen & LogReg & 0.782 & 0.771 & 0.565 & 0.374 & 0.176 & 0.06 & 0.034 \\
      & MLP1 &  0.758 & 0.752 & 0.53 & 0.337 & 0.164 & 0.07 & 0.044 \\
      & MLP4 & 0.752 & 0.758 & 0.521 & 0.312 & 0.003 & -0.478 & -0.685 \\
      & mttri & 0.821 & 0.832 & 0.642 & 0.489 & 0.235 & -0.192 & -0.384 \\
      & T5 & 0.777 & 0.768 & 0.555 & 0.332 & -0.113 & -1.004 & -1.449 \\
  	& SieBERT & \textbf{0.865} & \textbf{0.859} & \textbf{0.73} & 0.595 & 0.328 & -0.195 & -0.454  \\
  	& GPT-3 & 0.853 & 0.851 & 0.706 & \textbf{0.599} & \textbf{0.464} & \textbf{0.308} & \textbf{0.251}  \\
      \hline
    \end{tabular}%
    }
    \caption{Comparison between accuracy, F1-score and value in an OOD setting. LogRef, MLP1, ML4 and mttri are trained to perform domain adaptation and thus operate in a OOD setting, while transformer models (T5, SieBERT, GPT-3) are pre-trained on large corpora and thus likely operate in-distribution. For each dataset and metric, the best performance is highlighted in bold.}  
    \label{tab:ood}
    \end{sc}
  \end{center}
\end{table}

Table~\ref{tab:ood} reports the results of all models on the MDS tasks. As expected, large pre-trained language models tend to perform well across the board. This can be due to two reasons (besides the models being very powerful): (i) we know that very large models with very large train datasets are reasonably well calibrated  (e.g. \cite{Jiang2021}), and (ii) when the training data is so large, fewer examples are out of distribution in terms of language. For example, GPT-3 \cite{GPT3} is trained on about 45TB of text data from various datasets, and the vocabulary of the MDS datasets is most likely already present in its training set.

Notice however that even for these models, accuracy is a poor proxy of value when $k$ is large. Indeed, SieBERT slightly outperforms GPT-3 in terms of both accuracy and F1 in all tasks. However, the situation is reversed for large values of $k$, with SieBERT reaching negative values in most cases, most likely because of a poorer calibration with respect to GPT-3. Finally, simple linear models occasionally outperform these powerful (and very expensive to employ) large-language models for the largest values of $k$, again confirming the importance of value in determining the most appropriate model for the situation at hand.

\subsection{Key Takeaways for AI-Assisted Decision-Making} \label{insights}

While our experiments are conducted in a controlled setting, they are designed to reflect realistic decision-making scenarios relevant to enterprise environments, such as those in ServiceNow. Consider, for instance, an application that assesses or explains risk levels (e.g., the risk of applying a system patch). The utility of AI outputs in this context depends on the nature of potential errors:

\begin{itemize}
    \item \textit{Correct Assessments:} AI provides accurate risk evaluations, aiding decision-makers (e.g., Change Approvers) in making informed choices.
    \item \textit{Low-Value Outputs:} AI offers insights that, while accurate, do not significantly aid decision-making.
    \item \textit{Erroneous Assessments:} AI produces misleading risk evaluations (e.g., downplaying a high-risk change), potentially leading to poor decisions.
\end{itemize}

To mitigate the impact of errors, we apply a cost-based evaluation framework that assigns heavily negative weights to erroneous assessments—especially those that underestimate risks—relative to the positive weights for correct assessments. This reflects a deliberate design principle: it is preferable to provide no assistance than to offer misleading guidance.

We determine whether to deploy a model by setting penalties such that a positive overall score indicates a net beneficial impact. While this introduces cost as an additional parameter, it aligns with standard model evaluation practices, where accuracy and utility thresholds guide deployment decisions. Importantly, this framework prioritizes the decision-maker’s needs, resulting in more instances of model rejection (when thresholds are unmet) rather than erroneous inferences.

Notably, we have yet to encounter a use case where correct and erroneous assessments are assigned equal absolute weights by product managers. Similarly, non-inference (a model opting out) is rarely considered as detrimental as providing incorrect guidance. These observations suggest that our evaluation framework aligns with real-world utility considerations.

\section{Limitations and Conclusion} \label{conclusion}
In this paper, we investigated whether (i) model accuracy or F1-score serves as a reliable proxy for evaluating the true value of ML models, (ii) cost-sensitive error provides a meaningful measure of model value in cost-sensitive scenarios, (iii) calibration influences the value of ML models, and (iv) predictions in out-of-distribution settings impact model value. Our study focused on binary and multi-class classification tasks, employing various models under different cost settings. The findings revealed that (i) accuracy and F1-score are poor indicators of model value, (ii) cost-sensitive error is also an inadequate measure of model value, (iii) poor calibration significantly diminishes model value, and (iv) operating in out-of-distribution settings considerably undermines model value.

The takeaway from our experiments is that using accuracy-oriented metrics (that is, metrics that assume models are applied without rejection) is as a minimum a risky proposition - and this is true even for models widely acknowledged as ``leaders''. We should always assess models over a range of cost factors, and at least for reasonable cost factors we expect based on the set of application use cases we are targeting. $k=0$ (accuracy) is almost never a reasonable one.
We also saw how applying models without thresholding can lead to a negative value, and that threshold tuning seems to perform better than calibration. We also hypothesize and have obtained some support for identifying complexity and out-of-distribution as factors that may lead to rapid model quality degradation for higher cost factors.

This being said, we see this work more as providing evidence of a problem and outlining the research needs: more studies (especially with large models and in vs out of distribution datasets) are needed to validate the hypothesis and a deeper understanding of how calibration, confidence distribution, and size of validation set affect model value.


\backmatter








\section*{Declarations}

\noindent \textbf{Funding.} Funded by the European Union. Views and opinions expressed are however those of the author(s) only and do not necessarily reflect those of the European Union or the European Health and Digital Executive Agency (HaDEA). Neither the European Union nor the granting authority can be held responsible for them. Grant Agreement no. 101120763 - TANGO. Grant Agreement no. 952215 - TAILOR. The work of Burcu Sayin was partially supported by the project AI@Trento (FBK-Unitn). AP also acknowledges the support of the MUR PNRR project FAIR - Future AI Research (PE00000013) funded by the NextGenerationEU.
\newline

\noindent \textbf{Competing interests.} The authors declare that the research was conducted in the absence of any commercial or financial relationships that could be construed as a potential conflict of interest.
\newline

\noindent \textbf{Author contributions}.

\begin{itemize}
    \item Burcu Sayin: Conceptualization, Implementation, Experimental Evaluation, Writing – original draft, Writing – review \& editing. 

    \item Jie Yang: Conceptualization, Supervision, Writing – review \& editing.

    \item Xinyue Chen: Experimental Evaluation, Writing – original draft.

    \item Andrea Passerini: Conceptualization, Funding acquisition, Supervision, Writing – original draft, Writing – review \& editing. 

    \item Fabio Casati: Conceptualization, Funding acquisition, Supervision, Writing – original draft, Writing – review \& editing. 
\end{itemize}

\bibliography{main}

\end{document}